\documentclass[journal, letterpaper]{IEEEtran}

\usepackage{url}        

\usepackage{amsmath}

\usepackage{textgreek}	
\usepackage{listings}
\usepackage{csvsimple}
\usepackage{longtable}
\usepackage{graphicx}
\usepackage{amsmath}
\usepackage{amssymb}
\usepackage{booktabs}
\usepackage{pifont}
\usepackage{subcaption}
\begin{document}

        \title{BEV-LGKD: A Unified LiDAR-Guided Knowledge Distillation Framework for BEV 3D Object Detection}
        \author{
            \IEEEauthorblockN{Jianing Li$^{a}$, Ming Lu$^{c}$, Jiaming Liu$^b$, Yandong Guo$^d$,Li Du$^a$,Shanghang Zhang$^b$}\\
            \IEEEauthorblockA{$^a$ Nanjing University}
            \IEEEauthorblockA{$^b$ Peking University}
            \IEEEauthorblockA{$^c$ Intel Lab China}
            \IEEEauthorblockA{$^d$ Beijing University of Posts and Telecommunications}
        }
        \maketitle

\begin{abstract}
Recently, Bird's-Eye-View (BEV) representation has gained increasing attention in multi-view 3D object detection, which has demonstrated promising applications in autonomous driving. Although multi-view camera systems can be deployed at low cost, the lack of depth information makes current approaches adopt large models for good performance. Therefore, it is essential to improve the efficiency of BEV 3D object detection. Knowledge Distillation (KD) is one of the most practical techniques to train efficient yet accurate models. However, BEV KD is still under-explored to the best of our knowledge. Different from image classification tasks, BEV 3D object detection approaches are more complicated and consist of several components. In this paper, we propose a unified framework named BEV-LGKD to transfer the knowledge in the teacher-student manner. However, directly applying the teacher-student paradigm to BEV features fails to achieve satisfying results due to heavy background information in RGB cameras. To solve this problem, we propose to leverage the localization advantage of LiDAR points. Specifically, we transform the LiDAR points to BEV space and generate the foreground mask and view-dependent mask for the teacher-student paradigm. It is to be noted that our method only uses LiDAR points to guide the KD between RGB models. As the quality of depth estimation is crucial for BEV perception, we further introduce depth distillation to our framework. Our unified framework is simple yet effective and achieves a significant performance boost. Code will be released.
\end{abstract}

\begin{figure}[t]
\includegraphics[width=8.99cm]{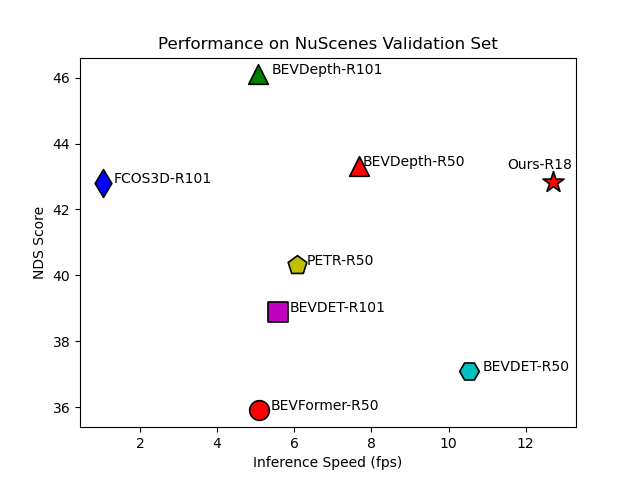}
\centering
\caption{Inference speed of different methods on nuScenes val set. We test the inference speed on a single NVIDIA V100-32G GPU with the batch size of 1.}
\label{fig:intro}
\end{figure}

\section{Introduction}

3D object detection is an essential computer vision technique with wide application scenarios such as autonomous driving. Recently, multi-view 3D object detection has gained increasing attention thanks to significant improvements in the results of Bird's-Eye-View (BEV) perception. As a common representation of surrounding scene, BEV can clearly present the location and scale of objects. Compared with multi-modal systems, multi-view camera systems can be deployed at low cost, while the lack of depth information makes current approaches adopt large models for good performance. Therefore, it is crucial to improve the efficiency of current approaches for practical deployment on vehicles.

Knowledge Distillation (KD) is an effective method to train efficient yet accurate neural networks and has been extensively studied in fundamental tasks like image classification. \cite{hinton2015distilling} presents the well-known teacher-student paradigm, which forces the logits of a smaller network (student) to match the logits predicted by a large network (teacher). Many subsequent works follow this paradigm but match the hidden layer features as extra knowledge. However, different from image classification, BEV 3D object detection approaches are more complicated and consist of several components. Firstly, it is a multi-view system in which each camera covers a certain field of view. Existing methods construct the BEV feature by aggregating the camera features, which contain heavy background information \cite{huang2021bevdet, li2022bevdepth, liu2022petr}. Secondly, as pointed out by \cite{li2022bevdepth}, the quality of intermediate depth is the key to improving BEV 3D object detection.

To solve the problems, we propose a unified framework named BEV-LGKD based on the teacher-student paradigm. Compared with cameras, LiDAR points can capture precise 3D spatial information of foreground objects. Although many approaches have been proposed to explore the incorporation of LiDAR points for BEV 3D object detection, it should be noted that our method only uses LiDAR points to guide the KD between RGB models. The additional computational cost of our method is transforming the LiDAR points to BEV space, which is almost negligible compared with network training. We leverage the BEV LiDAR points and the camera parameters to obtain the foreground mask and view-dependent mask for teacher-student paradigm. The foreground mask can select the most informative regions for feature matching. The view-dependent mask exploits the characteristics of each view's feature. Therefore, the proposed BEV-LGKD can outperform other feature-based distillation methods and is especially suitable for BEV 3D object detection.
We further observe that the quality of intermediate depth can be improved with a large depth estimation network. Therefore, we design a novel depth distillation loss to further improve the 3D object detection performance. Our unified framework is simple yet effective and achieve significant performance gains. 

The contributions can be concluded as follows:
\begin{itemize}
\item We propose a unified BEV KD framework named BEV-LGKD that effectively leverages the LiDAR points to select informative foreground regions for BEV feature matching.

\item We introduce a novel depth distillation loss to our framework, helping the student model obtain more accurate depth estimation results, further improving the whole KD performance.

\item We conduct extensive experiments to evaluate the advantages of the proposed framework against other feature-based distillation methods.
\end{itemize}

\section{Related work}
 
{\bf Knowledge distillation} Knowledge Distillation (KD) is invented to transfer useful knowledge from a large teacher model to a small student model. Early works concentrate on matching the logits of image classification via adjusting a temperature coefficient during training \cite{hinton2015distilling, gou2021knowledge}. The following methods extend the teacher-student paradigm to dense prediction tasks, demonstrating its effectiveness  \cite{chen2017learning, liu2020structured,shu2021channel}. As for the object detection task, \cite{chen2017learning} proves the importance of feature learning during KD process. \cite{guo2021distilling, zhang2020improve} find that features of foreground and background regions contribute differently to the KD process. \cite{kang2021instance} proposes an instance-conditional framework to improve the KD performance. KD has been demonstrated the effectiveness on many vision tasks such as optical flow prediction\cite{wang2021end}, depth estimation\cite{wang2021knowledge, pilzer2019refine, liu2020structured} and semantic segmentation\cite{liu2020structured, he2019knowledge, ji2022structural}. Although there are some KD methods for object detection, BEV KD is still under-explored to the best of our knowledge. The most related method is one ICLR 2023 blind submission, which studies the problem of cross-modal distillation. They match the features of LiDAR model and RGB model, while our method only uses LiDAR points to guide the KD between RGB models. The additional computational cost of our method is transforming the LiDAR points to BEV space, which is almost negligible compared with network training. 

{\bf Vision-centric 3D object detection} Vision-centric 3D object detection is useful for applications like autonomous driving since camera systems can be deployed at a low cost. FCOS3D \cite{wang2021fcos3d} first decouples 3D targets into 2D and 3D attributes, and then predicts 3D objects by projecting the 3D center from 2D feature planes. PGD \cite{wang2022probabilistic} analyzes the advantage of applying depth distribution to 3D monocular object detection. DETR3D \cite{wang2022detr3d} follows the seminal work of DETR \cite{carion2020end}, which uses object queries to match the position and class information of the instances. Recently, Bird's-Eye-View (BEV), as a unified representation of surrounding views, has attracted increasing attention from the community. BEVDet \cite{huang2021bevdet} utilizes the LSS operation \cite{philion2020lift} to transform 2D image features to 3D BEV feature. PersDet \cite{zhou2022persdet} improves the BEV feature generation and proposes the perspective BEV detection framework. PETR \cite{liu2022petr} introduces 3D positional embedding to obtain the 3D position-aware features. BEVDet4D \cite{huang2022bevdet4d} and PETRv2 \cite{liu2022petrv2} both fuse the multi-frame features using a spatial-temporal alignment operation and achieve significant performance improvement. BEVFormer \cite{li2022bevformer} exploits the spatial and temporal cross-attention mechanism to query a BEV feature according to its position in BEV space. BEVDepth \cite{li2022bevdepth} finds that accurate depth estimation is essential for accurate BEV 3D object detection. BEVStereo \cite{li2022bevstereo} proposes to improve the depth estimation of camera-based systems by leveraging the temporal multi-view stereo (MVS) technology. They further design an iterative algorithm to update more valuable candidates, making it adaptive to moving candidates. STS \cite{wang2022sts} also tries to improve the depth estimation. They propose a novel technique that leverages the geometry correspondence to facilitate accurate depth learning. Although there are plenty of methods for BEV 3D object detection, BEV KD is under-explored as far as we know. 

\begin{figure*}[t]
	\includegraphics[width=17.2cm]{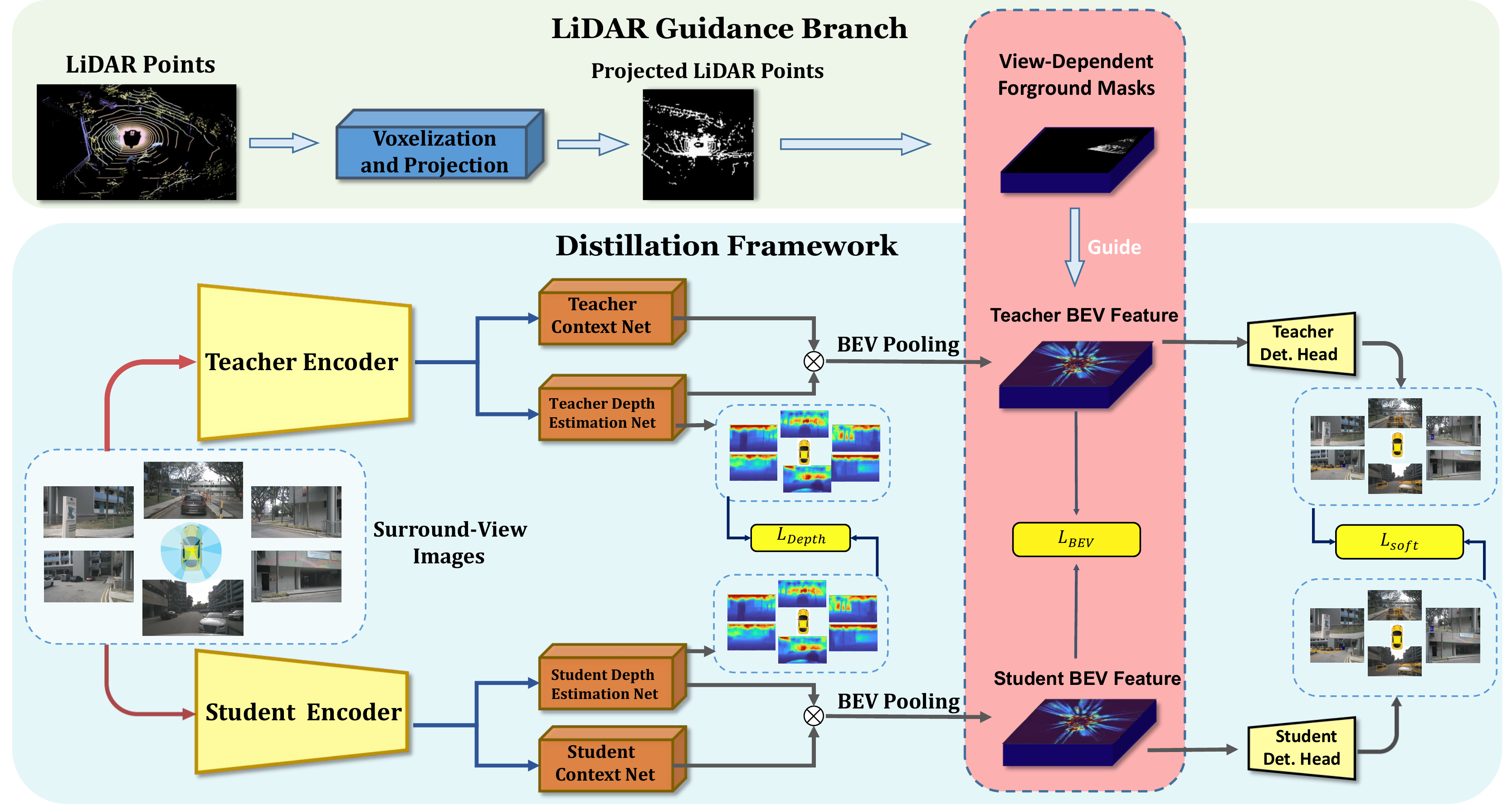}
	\centering
	\caption{\textbf{The framework of the proposed LGKD method}. The framework of LGKD follows the teacher-student paradigm with RGB images as the inputs. It consists of three distillation components: LiDAR-Guided BEV Distillation, Depth Distillation and Soft-label Distillation.}
	\label{fig:all}
\end{figure*}

{\bf Depth estimation} Since depth estimation is essential for vision-centric 3D object detection, we introduce related methods on depth estimation. Monocular and stereo are two most typical ways of depth estimation. Monocular methods \cite{eigen2014depth, fu2018deep, godard2019digging, guizilini20203d, bhat2021adabins, ranftl2021vision} generally build an encoder-decoder architecture to regress the depth map from contextual features. Previous methods tend to either use a regression head to predict dense depth map \cite{eigen2014depth, poggi2020uncertainty, ranftl2021vision} or use a classification head to predict a distribution along the depth range \cite{fu2018deep, bhat2021adabins}. Compared with monocular methods, stereo methods usually construct a cost volume to regress disparities based on photometric consistency \cite{khamis2018stereonet, chang2018pyramid, zhang2019ga, guo2019group, shen2021cfnet, peng2022rethinking}. We conduct analysis to the depth estimation component of BEV 3D object detection and find that KD between depth estimation models can further improve the performance of BEV 3D object detection.


\section{Method}
This section provides a detailed introduction to the proposed BEV LiDAR-Guided Knowledge Distillation (BEV-LGKD) framework. Our goal is to transfer the knowledge from a large RGB teacher model to a small RGB student model. The overall framework is illustrated by Fig. \ref{fig:all}. We use BEVDepth \cite{li2022bevdepth} as the baseline since it is a simple yet effective method. Our framework consists of three components: LiDAR-Guided BEV Distillation, Depth Distillation and Soft-label Distillation. We will introduce each component in the following sections.

Given an input multi-view image ${I_k} \in {R^{3 \times H \times W}}$, BEVDepth adopts a shared backbone model to extract the feature ${F_k} \in {R^{C \times {H_f} \times {W_f}}}$, where k is the index of the camera. They also predict the depth distribution map for each input image ${D_k} \in {R^{D \times {H_f} \times {W_f}}}$. Then they project the camera features to viewing frustum ${V_k} \in {R^{C \times D \times {H_f} \times {W_f}}}$ and sum up the frustum features falling into the same flattened BEV grid $B \in {R^{C \times {H_e} \times {W_e}}}$. Finally, the task-specific heads are applied to the BEV feature. We first introduce how to distill the knowledge between BEV features.

\begin{figure*}[t]
\includegraphics[width=17.2cm]{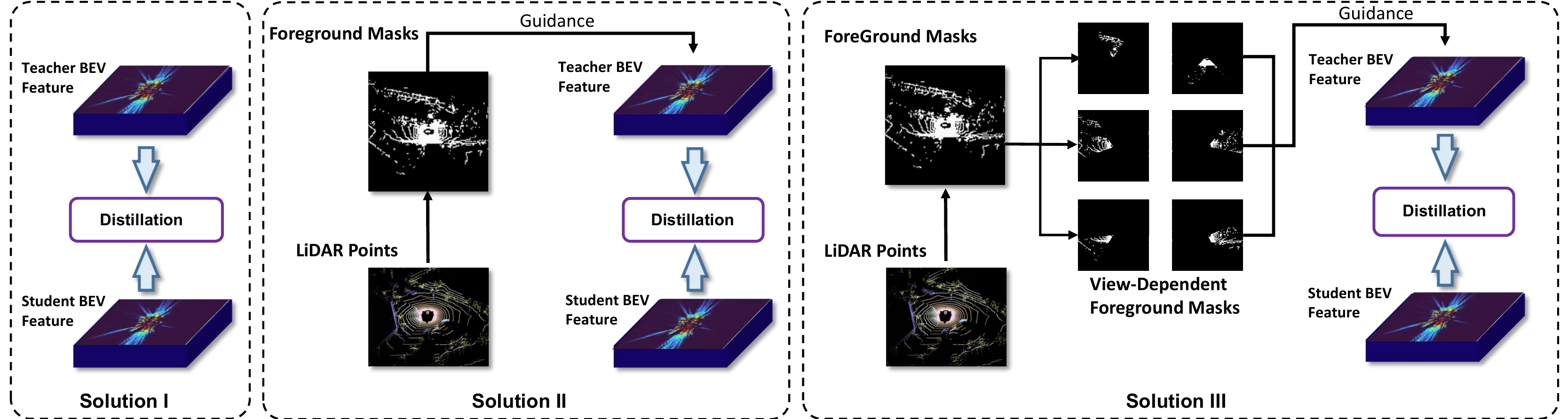}
\centering
\caption{\textbf{Different solutions of BEV feature distillation}. Solution I is the direct distillation in the feature level. Solution II is the feature distillation guided by LiDAR foreground mask. Solution III is the feature distillation guided by both LiDAR foreground and view-dependent masks.}
\label{fig:bevkd}
\end{figure*}

\subsection{LiDAR-guided BEV distillation}

BEV is a unified representation of surrounding views, thus BEV 3D object detection has become prevailing in multi-view 3D object detection recently. In this paper, we explore the BEV feature KD to improve the performance of multi-view 3D object detection. According to recent 2D and 3D object detection works \cite{yin2022fi, chen2021dual, pan2021model, yin2021center}, features close to object centers typically contain more useful information while features in background regions are less useful for KD. Therefore, directly applying the teacher-student KD paradigm to BEV feature fails to achieve satisfying results due to heavy background in BEV features. This is because BEV feature is aggregated from the multi-view RGB features, which contain heavy background information. To solve these limitations, we adopt the strong localization ability of the LiDAR sensor to help the KD between RGB models. LiDAR points can provide accurate location information of foreground objects. As illustrated by Fig. \ref{fig:bevkd}, we propose to adopt LiDAR points to distill useful information from teacher model to student model.

To be more specific, we first transform the LiDAR points to the BEV space and then voxelize \cite{zhou2018voxelnet} the projected LiDAR points to form a binary mask $M \in {R^{1 \times {H_e} \times {W_e}}}$. The binary mask indicates the occupancy status of flattened BEV grid. We further perform a Gaussian smoothness to extend the localization information from isolated positions ${M_g} = {g_\sigma }(M)$. Since BEV feature is aggregated from multiple views, we split the foreground mask into multiple overlapped masks $\left\{ {M_g^k} \right\}$ according to the view-dependent masks. The view-dependent masks are calculated by the camera field of view. Given the BEV features of teacher ${B_t} \in {R^{C \times {H_e} \times {W_e}}}$ and student ${B_s} \in {R^{C \times {H_e} \times {W_e}}}$. Our LiDAR-Guided BEV distillation loss is defined as follows.
\begin{equation}
{L_{bev}} = \sum\limits_{k = 1}^K {{\ell _2}(M_g^k \odot {B_t},M_g^k \odot {B_s})} 
\end{equation}
where $\ell_{2}$ is the L2 distance between masked BEV features and K is the number of cameras.

\subsection{Depth distillation}

As pointed out by BEVDepth, accurate depth estimation is essential for the performance of BEV 3D object detection \cite{li2022bevdepth}. Therefore, we design two kinds of losses for depth distillation to further improve the KD performance. The coarse depth is defined as the distribution of predefined depths. The fine depth is defined as the regressed dense depth values.

\textbf{Coarse depth loss} Since the coarse depth is supervised by the Binary Cross-Entropy (BCE) loss as the classification tasks, we follow the well-know method \cite{hinton2015distilling} and define the coarse depth loss as follows.
\begin{equation}
{L_{cd}} = T^{2} \sum\limits_{k = 1}^K {{\ell _{ce}}(D_k^s/T,D_k^t/T)}
\end{equation}
where $T$ is the temperature coefficient, and $\ell _{ce}$ is the Cross-Entropy loss.

\textbf{Fine depth loss} In order to enhance the depth distillation, we add a decoder $\phi$ to the context feature of BEVDepth \cite{li2022bevdepth} to regress the fine depth ${\widetilde D_k} = \phi ({F_k}) \in {R^{{H_f} \times {W_f}}}$. We use different structures for the decoders of teacher model and student model. For the fine depth, we use the L2 distance between the depth predictions of teacher and student:
\begin{equation}
{L_{fd}} = \sum\limits_{k = 1}^K {{\ell _2}(\widetilde D_k^s,\widetilde D_k^t)}
\end{equation}

The final loss for depth distillation is defined as:
\begin{equation}
{L_d} = {L_{cd}} + \alpha {L_{fd}}
\end{equation}
where $\alpha$ is the weight to balance coarse depth loss and fine depth loss. The depth distillation is illustrated in Fig. \ref{fig:all}.

\begin{table*}[t]
\small
  \centering
  \caption{3D object detection results of different methods on nuScenes val set. We list the results of the state-of-the-art methods to make direct comparisons. We compare the results of DETR3D \cite{wang2022detr3d}, FCOS3D \cite{wang2021fcos3d}, BEVDET \cite{huang2021bevdet}, PETR \cite{liu2022petr}, PGD \cite{wang2022probabilistic}, CenterNet \cite{duan2019centernet}, PersDet \cite{zhou2022persdet} and BEVDepth \cite{li2022bevdepth}. "C" in the modality column refers to the camera-only methods, and "L" refers to the Lidar-only methods. As can be seen, our method with a light-weight backbone (ResNet-18) can achieve competitive performance compared with BEVDepth (ResNet-50). Besides, BEV-LGKD achieves a significant performance boost compared with baseline.} 
    \setlength{\tabcolsep}{2.0mm}{
    \begin{tabular}{c|ccc|c|cccccc}
    \toprule
    \toprule
    Method & Image Size & Backbone & Modality & \multicolumn{1}{l}{NDS ↑} & \multicolumn{1}{l}{mAP ↑} & \multicolumn{1}{l}{mATE ↓} & \multicolumn{1}{l}{mASE↓}  & \multicolumn{1}{l}{mAOE↓} & \multicolumn{1}{l}{mAVE↓} & \multicolumn{1}{l}{mAAE↓}\\
    \midrule \midrule
    PointPillars & - & - & L & 0.597 & 0.487 &  0.315 & 0.260 & 0.368& 0.323& 0.203 \\
    \midrule
    CenterNet & - & DLA & C & 0.328 &  0.306 & 0.716 &  0.264 &0.609 & 1.426 & 0.658 \\
    FCOS3D & $900 \times1600$ & ResNet-101 & C & 0.415 & 0.343 &  0.725 & 0.263 &  0.422& 1.292 &0.153 \\
    PGD & $900 \times1600$ & ResNet-101 & C & 0.428 & 0.369 & 0.683 & 0.260 & 0.439& 1.268& 0.185 \\
    \midrule
    BEVDet & $384\times1056$ & ResNet-101 & C & 0.389 &  0.317 &  0.704 & 0.273 & 0.531 &  0.940 & 0.250 \\
    BEVDet & $512\times1408$ & Swin-T & C &  0.417 & 0.349 & 0.637 & 0.269 & 0.490 & 0.914 &  0.268 \\
    PersDet & $512\times1408$ & ResNet-50 & C &  0.408 & 0.346  & 0.660 & 0.279 & 0.540 &  0.964 &   0.207  \\    
    BEVDepth & $256\times708$ & ResNet-50 & C &  0.435 & 0.330  & 0.702 & 0.280 & 0.535 &  0.553 &   0.227  \\    
    \midrule
    DETR3D & $900\times1600$ & ResNet-101 & C & 0.374  & 0.303 &  0.860 & 0.278 & 0.437 & 0.967 &  0.235 \\
    PETR & $512\times1408$ & ResNet-101 & C &  0.421 & 0.351 & 0.710 & 0.270 & 0.490 & 0.885 &  0.224 \\
    CrossDTR & $512\times1408$ & ResNet-101 & C &  0.426 & 0.370 & 0.773 & 0.269 & 0.482 & 0.866 &  0.203 \\
    BEVFormer-T & $900\times1600 $ & ResNet-50 & C &  0.359 & 0.357 & 0.899 & 0.294 & 0.655 & 0.657 &  0.216 \\
    \midrule
    Ours (baseline) & $256 \times708$ & \textbf{ResNet-18} & C & 0.372 & 0.275 & 0.740 & 0.289 & 0.708& 0.689 & 0.229 \\
    
    Ours (LGKD)& $256 \times708$ & \textbf{ResNet-18} & C & 0.425 & 0.305 & 0.701 & 0.273 & 0.560& 0.500 & 0.215 \\

    \bottomrule
    \bottomrule
    \end{tabular}%
    }
  \label{tab:nu_val}%
\end{table*}%

\subsection{Soft-label distillation}
For the detection task, we utilize a CenterNet \cite{zhou2019objects} head to predict object locations and labels. Based on the BEV feature map $B \in {R^{C \times {H_e} \times {W_e}}}$, a heatmap would be regressed to represent the centers and categories in the BEV view. With a soft-label distillation loss $L_{soft} = \ell^{d}( O_s, O_t )$, the student model would have a faster convergence process under the guide of the teacher model.

\subsection{Loss functions}
In the phase of task learning, following BEVDepth \cite{li2022bevdepth}, we use the multi-task loss of 3D object detection and depth estimation. Since this is not our main contribution, we refer the readers to BEVDepth \cite{li2022bevdepth}. In the distillation phase, the distillation loss is the weighted sum of our three components:
\begin{equation}
L = {L_{soft}} + \beta {L_{bev}} + \gamma {L_d}
\end{equation}
the hyper-parameters are fixed during all our experiments.
\section{Experiments}
\subsection{Settings}

\textbf{Datasets} We use the nuScenes\cite{caesar2020nuscenes} dataset to evaluate the performance of our distillation framework. NuScenes contains 1k sequences, each of which is composed of six groups of surround-view camera images, one group of Lidar data and their sensor information. The camera images are collected with the resolution of $1600 \times 900$ at 12Hz and the LiDAR frequency for scanning is 20Hz. The dataset provides object annotations every 0.5 seconds, and the annotations include 3D bounding boxes for 10 classes \{Car, Truck, Bus, Trailer, Construction vehicle, Pedestrian, Motorcycle, Bicycle, Barrier,  Traffic cone \}. We follow the official split that uses 750, 150, 150 sequences as training, validation and testing set respectively. So totally we get 28130 batches of data for training, 6019 batches for validation, and 6008 batches for testing. 

\begin{table*}[t]
\small
  \centering
  \caption{3D object detection results of different methods on the nuScenes test set. We list the results of the state-of-the-art methods to make direct comparisons. We compare with the results of DETR3D \cite{wang2022detr3d}, FCOS3D \cite{wang2021fcos3d}, BEVDET \cite{huang2021bevdet}, PETR \cite{liu2022petr}, PGD \cite{wang2022probabilistic}, CenterNet \cite{duan2019centernet}, "C" in the modality column refers to the camera-only methods, and "L" refers to the Lidar-only methods. As can be seen, our method with a light-weight backbone is still very competitive and BEV-LGKD also achieve a significant performance boost, demonstrating the generalization ability of BEV-LGKD.} 
    \setlength{\tabcolsep}{2.0mm}{
    \begin{tabular}{c|ccc|c|cccccc}
    \toprule
    \toprule
    Method & Image Size & Backbone & Modality & \multicolumn{1}{l}{NDS ↑} & \multicolumn{1}{l}{mAP ↑} & \multicolumn{1}{l}{mATE ↓} & \multicolumn{1}{l}{mASE↓}  & \multicolumn{1}{l}{mAOE↓} & \multicolumn{1}{l}{mAVE↓} & \multicolumn{1}{l}{mAAE↓}\\
    \midrule \midrule
    CenterNet & - & ResNet-101 & C & 0.400 &  0.338 & 0.658 &  0.255 &0.629 & 1.629 & 0.142 \\
    FCOS3D & $900 \times1600$ & ResNet-101 & C & 0.428 & 0.358 &  0.690 & 0.249 &  0.452& 1.434 &0.124 \\
    PGD & $900 \times1600$ & ResNet-101 & C & 0.448 & 0.386 & 0.626 & 0.245 & 0.451 & 1.509& 0.127 \\
    \midrule
    BEVDet & $512\times1408$ & Swin-S & C &  0.463 & 0.398 & 0.556 & 0.239& 0.414& 0.101 & 0.153 \\
    \midrule
    DETR3D & $900\times1600$ & V2-99 & C & 0.479  & 0.412 &  0.641 & 0.255 & 0.394 & 0.845 &  0.133 \\
    PETR & $512\times1408$ & ResNet-101 & C & 0.455 & 0.391 & 0.647 &  0.251 &  0.433 &  0.933 & 0.143  \\
    \midrule
    Ours & $256 \times708$ & \textbf{ResNet-18} & C & 0.453 & 0.327 & 0.632 & 0.265 &0.524 & 0.520 & 0.163 \\
    
    \bottomrule
    \bottomrule
    \end{tabular}%
    }
  \label{tab:nu_test}%
\end{table*}%

\textbf{Evaluation metric} For the 3D object detection task, we use mean Average Precision(mAP) and Nuscenes Detection Score(NDS) as our main evaluation metrics. We also adopt other officially released metrics concluding Average Translation Error (ATE), Average Scale Error (ASE), Average Orientation Error (AOE), Average Velocity Error (AVE), and Average Attribute Error (AAE). Note that NDS is a weighted sum of mAP and other metric scores. For the depth estimation task, we use Abs Relative difference  (Abs\_Rel) and prediction accuracy threshold ($\delta < 1.25$) as our main evaluation metrics. The other adopted metrics are Squared Relative difference (Sq\_Rel), Relative Mean Square Error (RMSE) and RMSE\_log.

\textbf{Implementation details} For both teacher and student models, we follow the pipeline of BEVDepth \cite{li2022bevdepth} while add the FCN decoders for depth distillation. The teacher model adopts ResNet-101 \cite{he2016deep} or ResNet-50 as the heavy backbone, and the student model adopts ResNet-18 or MobileNetv2 as the light-weight backbone. The teacher model uses FPN with channel outputs \{160, 160, 160, 160\}, while the student model uses FPN with channel outputs \{128, 128, 128, 128\}. For BEV, we set the detection range to [-51.2m, 51.2m] along $X$ and $Y$ axis. The image scale for student model is $704 \times 256$. We adopt multi-frame fusion strategy that is proposed in BEVDepth \cite{li2022bevdepth}. We \textbf{do not} adopt CBGS \cite{zhu2019class} strategy in the experiments to gain extra improvements. In the optimization phase, we adopt Pytorch-lightning to compile the whole framework and use AdamW \{weight\_decay=1e-7\} as the optimizer. We train the teacher model for 25 epochs with batch-size of 5 on 8 Nvidia Tesla V100 GPUs, and the student model with batch-size of 8 on 8 Nvidia Tesla V100 GPUs. The distillation process takes 35 epochs to finish.
\subsection{Comparison with the SOTA baselines}
\textbf{3D Object detection results on nuScenes validation set} As shown in Tab. \ref{tab:nu_val}, we compare with the state-of-the-art methods on nuScenes val set. We only compare with methods based on RGB cameras since our method does not require RGB during inference. Since the camera-based methods lack depth information, a heavy backbone is usually required to achieve satisfying performance, leading to huge computational costs. In contrast, our method relies on a lightweight backbone and effectively transfers knowledge from a heavy backbone. Tab. \ref{tab:nu_val} shows that our method with ResNet-18 can outperform typical monocular 3D object detection methods PGD \cite{wang2022probabilistic}, FOCOS3D \cite{wang2021fcos3d} and CenterNet \cite{duan2019centernet} with ResNet-101. For a fair comparison, we show the results of BEVDepth \cite{li2022bevdepth}, which is also our baseline model. As can be seen, our method with ResNet-18 can achieve competitive performance with BEVDepth with ResNet-50. The gain of BEV-LGKD reaches {\bf 5.6\%} on NDS compared with baseline, demonstrating the effectiveness of our method.

\textbf{3D Object detection results on nuScenes test set} We also report the results on nuScenes test set in Tab. \ref{tab:nu_test}. As can be seen, we also obtain remarkable results compared with the state-of-the-art methods. The proposed BEV-LGKD framework also achieves a significant performance boost, demonstrating the generalization ability of BEV-LGKD.

\begin{table*}[t]
	\centering
	\caption{Depth estimation results of different methods on nuScenes validation set. We compare with the results of FSM \cite{guizilini2022full}, SurroundDepth \cite{wei2022surrounddepth}, Monodepth2 \cite{godard2019digging} . The depth estimation shares the backbone with 3D object detection. As can be seen, our method with ResNet-18 can outperform the results of Monodepth2 and SurroundDepth with ResNet-34 on the depth estimation task. The performance boost of BEV-LGKD is also significant. "S" refers to self-supervised learning methods and "F" refers to full-supervised learning methods. } 
	\setlength{\tabcolsep}{1.2mm}{
		\begin{tabular}{c|c|c|cccc|ccc}
			\toprule\toprule
			Method & Backbone & Type  &\multicolumn{1}{l}{Abs Rel ↓} & \multicolumn{1}{l}{Sq Rel ↓} & \multicolumn{1}{l}{RMSE ↓} & \multicolumn{1}{l}{RMSE log ↓} & \multicolumn{1}{l}{$\delta > {1.25}$ ↑} & \multicolumn{1}{l}{$\delta > {1.25}^2$ ↑} & \multicolumn{1}{l}{$\delta > {1.25}^3$ ↑} \\
			\midrule \midrule
			Monodepth2 & ResNet-34 &   S &0.303 & 2.675 & 9.210 & 0.437 & 0.506 & 0.764 & 0.881 \\

			FSM &  ResNet-34& S & 0.298 & 2.586 & 8.996 & 0.420 & 0.541 & 0.767 & 0.888 \\
			SurroudDepth &  ResNet-34&  S & 0.245 & 3.067 & 6.835 & 0.321 & 0.719 & 0.878 & 0.935 \\
			\midrule
			Ours(teacher) & ResNet-101& F & 0.147 & 1.055  & 4.595 & 0.219 &0.829 & 0.931 & 0.969 \\
			Ours(student) & ResNet-18& F &0.190& 1.411 & 5.447 & 0.263 & 0.761 & 0.896 & 0.957 \\
			Ours(base) & ResNet-18& F & 0.204 & 1.603 & 5.892 & 0.274 & 0.737 & 0.880 & 0.921 \\
			\bottomrule
			\bottomrule
		\end{tabular}%
	}
	\label{tab:depth}%
\end{table*}%

\begin{table}[b]
	\scriptsize
	\caption{Depth estimation results of individual camera on nuScenes validation set.} 
	\setlength{\tabcolsep}{1.2mm}{
		\begin{tabular}{c|cccccc|c}
			\hline
			\hline
			& & &  \multicolumn{3}{c}{\textbf{Abs-Rel.↓}} \\
			Method &  \multicolumn{1}{l}{F\_Left} & \multicolumn{1}{l}{Front} & \multicolumn{1}{l}{F\_Right} & \multicolumn{1}{l}{B\_Left} & \multicolumn{1}{l}{Back} & \multicolumn{1}{l}{B\_Right} & \multicolumn{1}{l}{Std.↓}\\
			\hline\hline
			Monodepth2  & 0.304 &0.214 & 0.388 & 0.314 & 0.304& 0.438 & 0.078  \\
			FSM   &  0.287   & 0.186 & 0.375 & 0.296  & 0.221 & 0.418 & 0.088 \\
			\hline
			Ours(teacher)   & 0.150 & 0.094 & 0.165 & 0.159 & 0.127 & 0.186 & 0.032 \\
			Ours(student)  & 0.195 & 0.123 & 0.214 &0.205 & 0.164 & 0.233 & 0.039 \\
			\hline\hline
		\end{tabular}%
	}
	\label{tab:depth_each}%
\end{table}%

\textbf{Depth estimation results on nuScenes val set} We also compare the depth estimation on nuScenes val set. We compare our method with four state-of-the-art methods including Monodepth2 \cite{godard2019digging}, FSM \cite{guizilini2022full} and SurroundDepth\cite{wei2022surrounddepth}. Monodepth2 is self-supervised monocular depth estimation method while FSM and SurroundDepth are self-supervised multi-view depth estimation methods. The sparse ground truth is projected from LiDAR points by homographic warping. For the evaluation metrics, we get an average of results from 6 surround cameras. As shown in Tab. \ref{tab:depth}, our depth estimation results also have advantages on many metrics. Compared with monocular estimation methods like MonoDepth2, we obtain 11.3\% decrease in Abs\_Rel metric and 1.36 in Sq\_Rel metric. We also outperform the multi-view methods like FSM. To make a more fair comparison, we calculate the depth estimation performance for individual cameras and compare it with Monodepth2 in Tab. \ref{tab:depth_each}. As can be seen, our method also outperforms the monocular methods and the gain of BEV-LGKD is significant.

\begin{figure}[htbp]
        \centering
        \subcaptionbox{RGB}{
        \includegraphics[width = .2\linewidth]{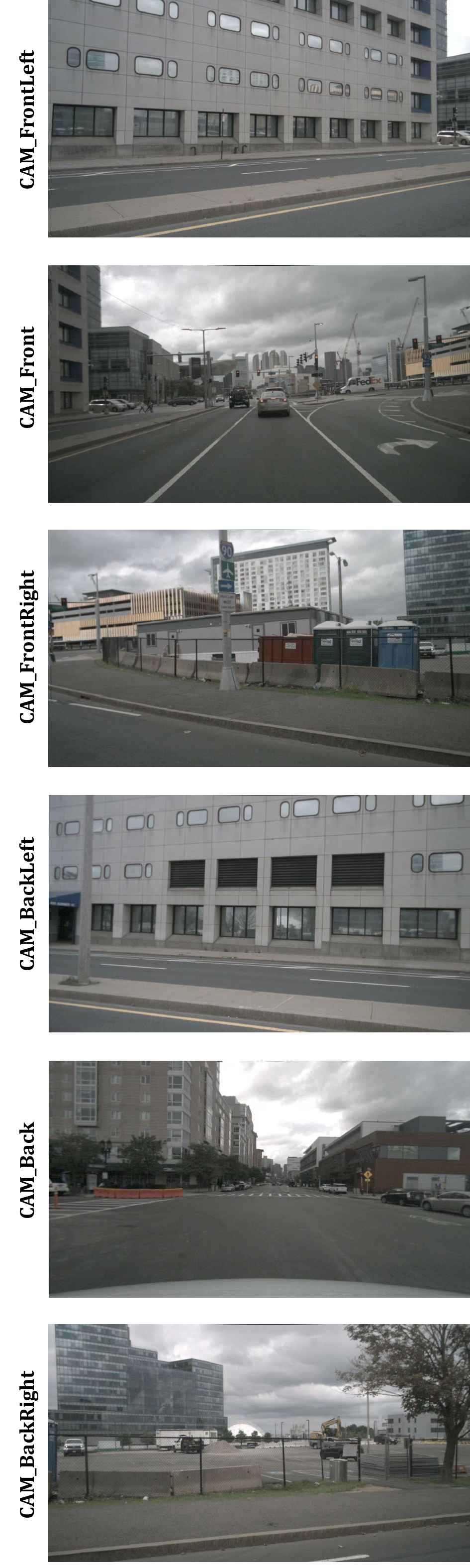}
    }
        \subcaptionbox{Base}{
        \includegraphics[width = .2\linewidth]{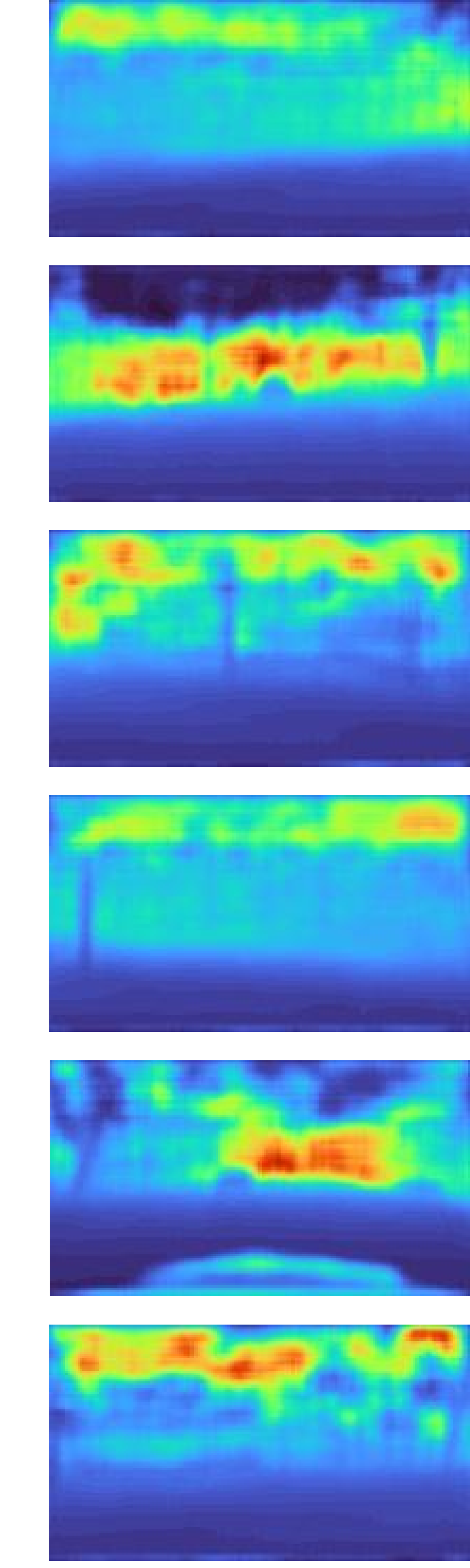}
    }
        \subcaptionbox{Teacher}{
        \centering
        \includegraphics[width = .2\linewidth]{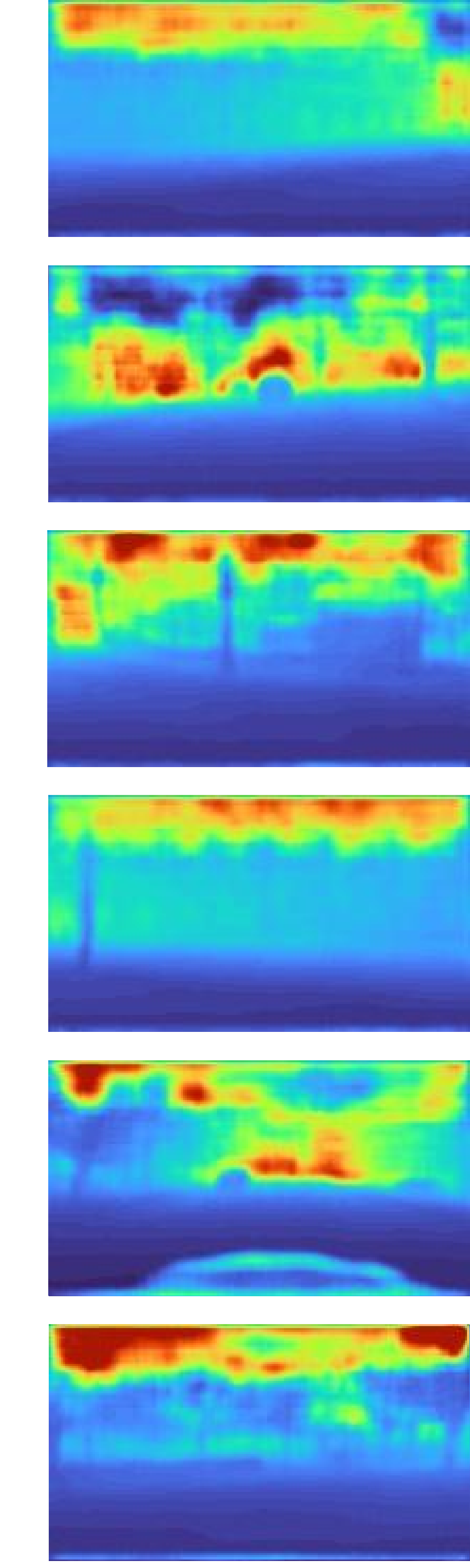}
    }
        \subcaptionbox{Distilled}{
        \includegraphics[width = .2\linewidth]{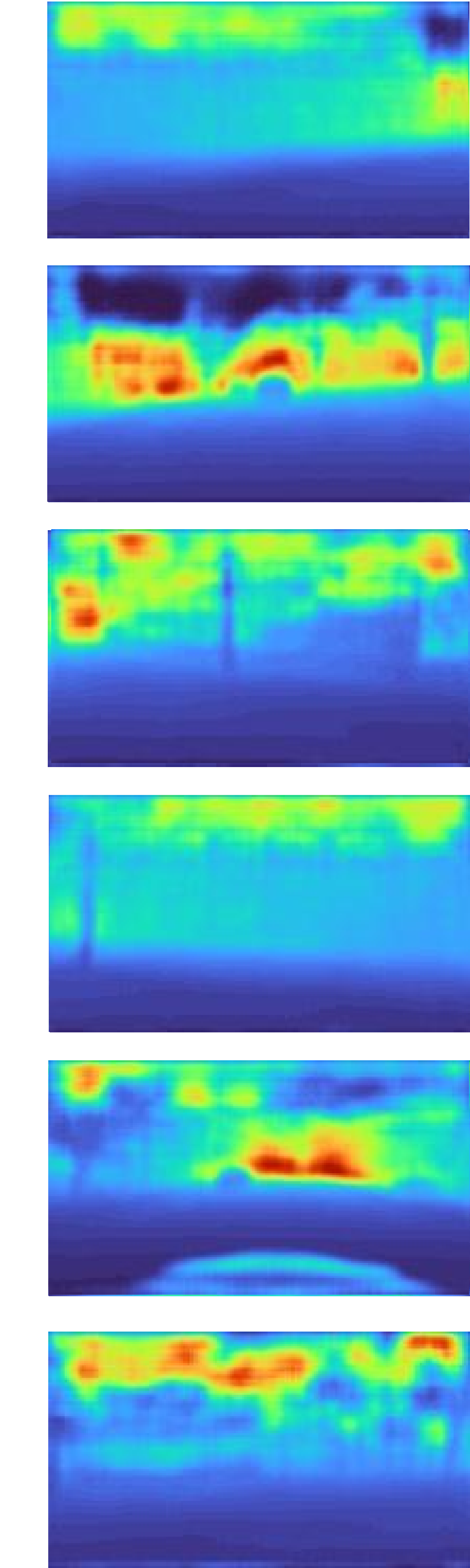}
    }
        \caption{Visualization of depth estimation results on nuScenes validation set. As shown in the figure, the depth distillation module improves the depth estimation performance with sharper edges and clearer shapes.}
        \label{fig:qual_depth}%
\end{figure}

\begin{figure*}[t]
	\centering
	\subcaptionbox{Base}{
		\includegraphics[width = .31\textwidth]{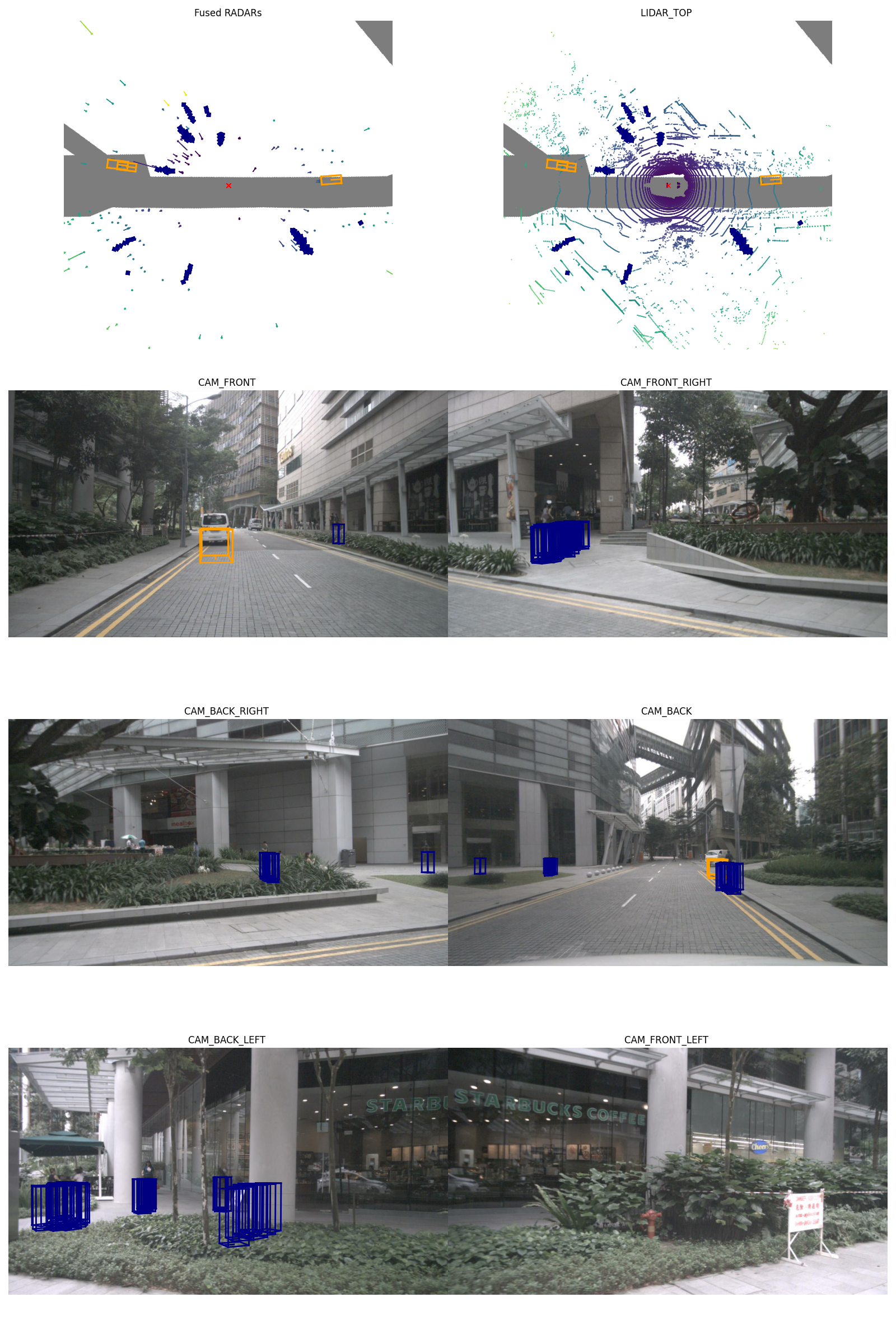}
	}
	\subcaptionbox{Distilled}{
		\centering
		\includegraphics[width = .31\textwidth]{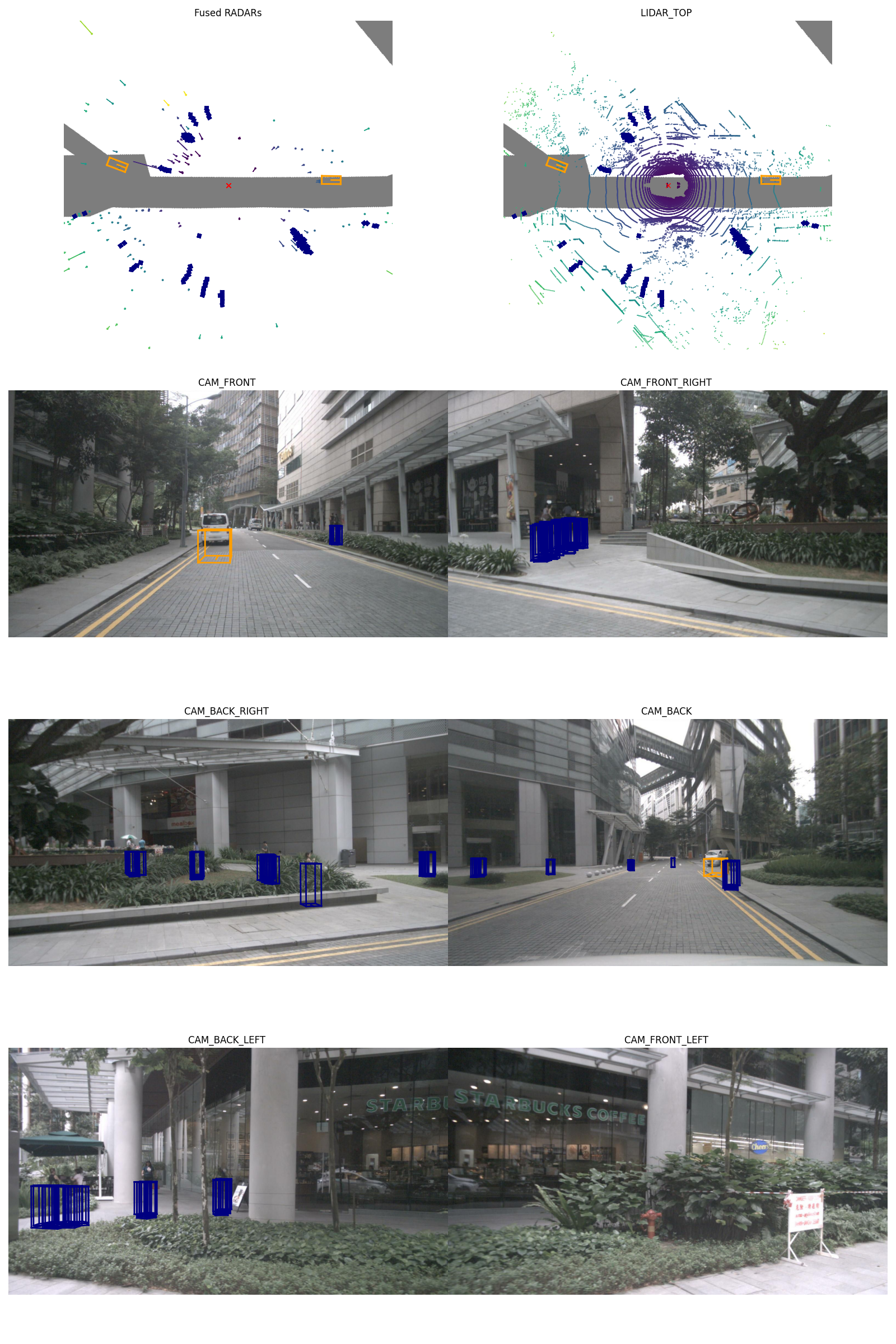}
	}
	\subcaptionbox{Ground-Truth}{
		\includegraphics[width = .31\textwidth]{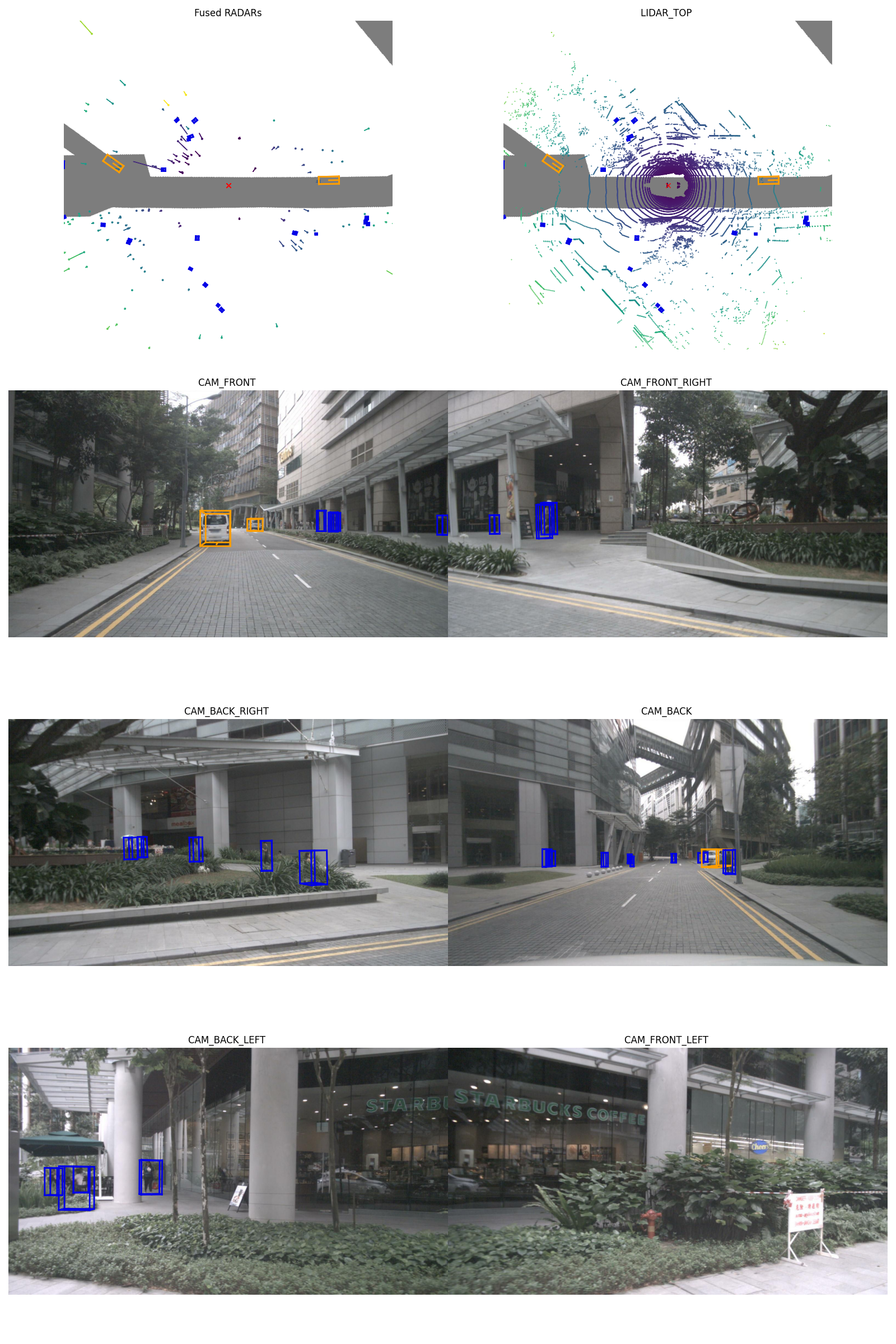}
	}
	\caption{Visualization results of 3D object detection on the nuScenes validation set. The student model (middle column) performs more accurate results than the un-distilled baseline (left column). We provide visualization for both surround camera views (bottom six images) and top-down LiDAR and RADAR views (top right and top left images).}
	\label{fig:qual_det}%
\end{figure*}

\textbf{Qualitative results} We show the qualitative results for both 3D object detection and depth estimation tasks. As can be seen from Fig. \ref{fig:qual_det} and Fig. \ref{fig:qual_depth}, the proposed BEV-LGKD improves the performance of both 3D object detection and depth estimation.

\begin{table}[h]
	\centering
	\begin{center}
		\caption{Ablation study of each component. As can be seen, the proposed three components contribute to the performance of 3D object detection. $L_{Task}^{\dagger}$ denotes all the task losses concluding detection and depth estimation, $L_{soft}$  is the soft label distillation loss, $L_{depth}$ is the depth distillation loss, and $ L_{bev} $ is the BEV feature distillation loss. }
		\setlength{\tabcolsep}{0.45mm}{
			\begin{tabular}{c|cccc|cc}
				\hline\hline
				Phase & $L_{Task}^{\dagger}$ & $L_{soft}$ & $L_{depth}$ & $ L_{bev} $ &  \multicolumn{1}{l}{NDS ↑}  & \multicolumn{1}{l}{mAP ↑} \\ 
				\hline
				\hline
				Base (R18) & \checkmark & - & - & - &  0.372 & 0.275 \\ 
				
			    $Exp_{1}$ & \checkmark & \checkmark & - & - &   0.391 & 0.290   \\ 
				$Exp_{2}$ & \checkmark & \checkmark & \checkmark & - &  0.407  & 0.294\\  
				$Exp_{3}$ & \checkmark & \checkmark & \checkmark & \checkmark &   0.425  & 0.305 \\ 
				$Exp_{4}$ & \checkmark & - & \checkmark & \checkmark &   0.416  & 0.301 \\ 
				\hline  
				Teacher (R101) & \checkmark & - & - & - & 0.471  & 0.350  \\ 
				\hline \hline
		\end{tabular}}
		\label{tab:each_comp}
	\end{center}
\end{table}

\begin{table}[b]
	\small
	\caption{Ablation Study of LiDAR Guidance} 
	\setlength{\tabcolsep}{1.2mm}{
		\begin{tabular}{c|c|c|ccccc}
			\hline
			\hline
			Method &  \multicolumn{1}{c}{NDS↑} & \multicolumn{1}{c}{mAP↑} & \multicolumn{1}{l}{mATE↓}  & \multicolumn{1}{l}{mAAE↓}\\
			\hline\hline
			Base-Direct  & 0.407(+ 3.5\%) & 0.294 (+ 1.5\%)& 0.720 & 0.225 \\
			Foreground  & 0.419(+ 4.7\%) & 0.304(+ 1.9\%) & 0.709  & 0.224 \\
			LGKD (Ours)  & 0.425(+ 5.3\%) & 0.305(+ 2.0\%) & 0.701  & 0.215 \\ 
			\hline\hline
		\end{tabular}%
	}
	\label{tab:lidar}%
\end{table}%

\begin{table}[b]
	\small
	\caption{Ablation Study of Other Distillation Methods} 
	\setlength{\tabcolsep}{1.2mm}{
		\begin{tabular}{c|c|c|ccccc}
			\hline
			\hline
			Method &  \multicolumn{1}{c}{NDS↑} & \multicolumn{1}{c}{mAP↑} & \multicolumn{1}{l}{mATE↓}  & \multicolumn{1}{l}{mAAE↓}\\
			\hline\hline
			FitNet  & 0.389 (+ 1.7\%) &0.283 (+ 0.8\%) & 0.742  & 0.228 \\
			LEKD   &  0.391(+ 1.9\%) & 0.288(+ 1.3\%) & 0.734  & 0.232 \\
			\hline
			LGKD(Ours)  & 0.428(+ 5.3\%) & 0.305(+ 3.0\%) & 0.701  & 0.215 \\
			LEKD+LGKD & 0.434(+ 6.2\%) & 0.310(+ 3.5\%) & 0.705  & 0.223 \\
			\hline\hline
		\end{tabular}%
	}
	\label{tab:feat_kd}%
\end{table}%

\begin{table}[h]
	\centering
	\begin{center}
		\caption{Ablation Study of Other Student Backbones.}
		\setlength{\tabcolsep}{0.45mm}{
			\begin{tabular}{c|c|ccc}
				\hline\hline
				Type & \multicolumn{1}{c}{Backbone} & \multicolumn{1}{c}{NDS} & \multicolumn{1}{c}{mAP} \\ 
				\hline
				\hline
				Teacher & Res-101 &  0.471  & 0.350 \\ 
				\hline
				$Base_{M}$  & Mob.Net-v2 &  0.237 & 0.142  \\ 
				$Student_{M}$ & Mob.Net-v2 & 0.275(+4.2\%) &   0.165 (+2.3\%) \\ 

				\hline\hline 
		\end{tabular}}
		\label{tab:mbnetv2_comp}
	\end{center}
\end{table}

\subsection{Ablation study}
\textbf{Ablation study of each component}
We conduct experiments to verify the effectiveness of the distillation modules for 3D object detection on nuScenes val set. We design three distillation components to transfer reliable information from teacher model including Lidar-guided BEV Distillation, Depth Distillation and Soft-label Distillation. To evaluate the effectiveness of each component, we design experiments to evaluate the performance of each component. As shown in Tab. \ref{tab:each_comp}, each component contributes to the distillation process in terms of different metrics. For the NDS score, the soft-label distillation contributes 1.9\% increase from 37.2\% to 39.1\%. The LiDAR-guided BEV distillation contributes most among the three components, improving the NDS score from \textbf{40.7\%} to \textbf{42.5\%} with a \textbf{1.8\%} gain. For the mAP score, the soft-label distillation brings 1.5\% improvements from 27.5\% to 29.0\% and the LiDAR-guided BEV distillation brings 1.1\% improvements from 29.4\% to 30.5\%. We also provide the performance of the teacher model and our method can narrow the performance gap between the teacher model and student model on the NDS score from 9.9\% to 4.6\%.

\textbf{Ablation study of LiDAR guidance}
Our method adopts the localization ability of LiDAR points to guide the knowledge distillation between RGB models. To validate the effectiveness of the proposed foreground and view-dependent masks, we further conduct an ablation study of LiDAR guidance. As shown in Tab. \ref{tab:lidar}, both the foreground mask and view-dependent mask contribute to the distillation process. Foreground masks contribute most to the performance gain since they can filter out the background information in BEV features.

\textbf{Ablation study of other student backbones}
In order to evaluate the generalization of our method, we choose MobileNet-v2 \cite{sandler2018mobilenetv2} as another light-weight backbone. We maintain the overall architecture and hyper-parameters for a fair evaluation. As shown in Tab. \ref{tab:mbnetv2_comp}, our method consistently improve the performance of 3D object detection using MobileNet-v2 as the light-weight backbone. The detailed record is listed in the supplementary file. 

\textbf{Ablation study of other distillation methods} We also compare with other feature distillation methods to demonstrate the advantage of the proposed framework. We choose the feature distillation method FitNet \cite{romero2014fitnets} and LEKD \cite{chen2017learning} as the alternatives of our LiDAR-guided BEV distillation component. For a fair comparison, we keep the depth distillation and soft-label distillation. As can be seen from Tab. \ref{tab:feat_kd}, the proposed LGKD method outperforms the feature distillation methods, demonstrating the advantage of BEV representation for multi-view perception tasks. We also conduct an experiment that combines feature distillation with BEV distillation and the additional margin is very small.
\section{Discussions and limitations}
In our work, we propose a novel distillation method to allow a lightweight model achieve remarkable performance for 3D object detection and depth estimation. However, our method still has some limitations. For example, we follow the pipeline of BEVDepth, which is sensitive to the accuracy of depth estimation. Besides, the performance on nuScenes test set is still not good enough compared with multi-modal methods \cite{liu2022bevfusion,yang2022deepinteraction}. We will make attempts to further improve the performance.

\section{Conclusion}
To summarize, we propose a novel and unified framework named BEV-LGKD for BEV 3D object detection. Our framework consists of three components including LiDAR-Guided BEV Distillation, Depth Distillation and Soft-label Distillation. We leverage the localization ability of LiDAR points to generate the foreground and view-dependent masks, which effectively filter out the background information in BEV features. Our method only uses LiDAR data to guide the KD training and does not require LiDAR sensors during inference. Since depth estimation is essential for camera-based systems, we further introduce the depth distillation component to the framework, significantly improving the qualities of 3D object detection and depth estimation. We demonstrate the effectiveness of our method through extensive experiments.

{\small
\bibliographystyle{ieee_fullname}
\bibliography{egbib}
}

\thispagestyle{empty}
\section{Appendix}
\appendix

\subsection{Additional related work}
\label{sec:A}
\textbf{LiDAR-based 3d detection} LiDAR-based 3d detectors usually project the points into voxels\cite{zhou2018voxelnet, mao2021voxel, deng2021voxel, shi2020pv, ye2020hvnet} or pillars\cite{lang2019pointpillars,wang2020pillar,nakamura2022robust} to form a regular representation due to the unordered nature of LiDAR data. Following these pre-processing stages, 2D and 3D convolutions are used to aggregate the 3D features, and project them to a Bird's-Eye-View space via a pooling operation, and form a \textbf{BEV feature}, where instances are clearly separated and easy to be detected. Similar to 2D detectors, some approaches like\cite{zhou2018voxelnet} operate 3D detection by groups of anchors while some \cite{yin2021center} operate 3D detection by the object centers. On the other hand, SECOND\cite{yan2018second} proposes a sparse computation flow to substantially accelerate 3D convolution because of the enormous computation and complexity for 3D convolutions.

\textbf{Lidar-image fusion for 3D object detection}
Recently, Lidar-image fusion approaches for 3D detection have gained much attention. Some approaches utilize the transformer architecture and make fusion on the object queries\cite{bai2022transfusion}, while others make fusion on the BEV features which are obtained from the camera branch and LiDAR branch\cite{liu2022bevfusion, wang2021pointaugmenting}. Both ways have demonstrated the advantage and effectiveness of fusing camera and LiDAR data when conducting 3D object detection.

\textbf{BEV segmentation}
BEV segmentation aims to segment useful regions such as lanes and vehicles in the BEV representation. LSS\cite{philion2020lift} first proposes to segment vehicles from an arbitrary number of cameras. Then, M2BEV\cite{xie2022m} operates joint learning of BEV segmentation and 3D object detection and BEVSegFormer\cite{peng2022bevsegformer} uses attention to query segmentation regions in BEV grids. CoBEVT\cite{xu2022cobevt} creatively proposes a multi-agent V2V(vehicle-to-vehicle) BEV segmentation framework and design a fused axial attention module to capture information interactions between views and agents. 

\textbf{BEVDistill}
We notice that a very recent work BEVDistill\cite{https://doi.org/10.48550/arxiv.2211.09386} propose a good method to conduct BEV feature distillation for multi-view 3D object detection. BEVDistill has achieved remarkable improvements on BEVFormer\cite{li2022bevformer}. The primary distinction between our work and BEVDistill is that we directly lead the distillation process using LiDAR points rather than using a LiDAR-based BEV feature to distill a camera-based BEV feature, as is adopted in BEVDistill. Both distillation methods have proven effective at 3D object detection, according to the improvement gaps between the teacher model and the student model.

\subsection{Task losses}
\textbf{3D Object detection}
Following BEVDet\cite{huang2021bevdet} and BEVDepth\cite{li2022bevdepth}, we adopt CenterNet\cite{duan2019centernet} head to regress object locations and categories. The detection head output a group of heatmaps to represent locations of different categories and assign scores to these keypoints. The loss function is:
\begin{equation}\nonumber
\tiny
    L_{c} = -\frac{1}{N} \sum_{c=1}\sum_{i=1}\sum_{j=1} \left\{
\begin{aligned}
(1-p_{cij})^{\alpha} log(p_{cij}) \quad if \quad y_{cij}=1 \\
(1-y_{cij})^{\beta}(p_{cij})^{\alpha}log(1-p_{cij}) \quad otherwise
\end{aligned}
\right.
\end{equation}
\textbf{Fine depth estimation}
We use Scale-Invariant Loss which is proposed in \cite{eigen2014depth} to regress dense depth by the supervision of LiDAR GT. In the following functions and metrics, $y$ and $y^{*}$ represent the predicted and ground-truth depth:
\begin{equation}
    L_{SIL} = \frac{1}{2n} \sum_{i=1}^{n} (log y_i - log y^{*}_i + f_{\alpha}(y_i, y^{*}_i))^{2}
\end{equation}
where $f_{\alpha}(y_i, y^{*}_i)=\frac{1}{n}\sum_i(log y_i - log y^{*}_i) $.

\subsection{Evaluation metrics}

\textbf{Depth estimation} Instead of setting any scale or offset to achieve a linear transformation, we use the original values of the sparse depth as the ground truth. And we decided to calculate metrics up to a distance of 80m.

\begin{itemize}
    \item Absolute relative error (AbsRel): $ \frac{1}{n} \sum_{i}^{n} \frac{|z_i - z_i^{*}|}{z^{*}} $

\item RMSE: $\sqrt{\frac{1}{N}\sum_{i=1}^{N}||D_i - \hat{D_i}||^{2}}$

\item RMSE\_log: $\sqrt{\frac{1}{N}\sum_{i=1}^{N}||log(\hat{D_i})-log(D_i)||^{2}}$

\item Sq\_Rel: $\frac{1}{N}\sum_{i=1}^{N}\frac{||D_i - \hat{D_i}||^{2}}{D_i}$

\item Accuracy: $\delta < \tau \% s.t. \quad max(\frac{\hat{D_i}}{D_i},\frac{D_i}{\hat{D_i}})=\delta<\tau$
\end{itemize}

\begin{table*}[t]
  \centering
  \caption{We include our Res-18 student model's per-class accuracy on the nuScenes validation set. According to our submitted paper, the mean AP increased by 3.5\%, from 27.5\% to 31.0\%, when compared to the basic model. Meanwhile, performances on three categories have gained increases by more than 5.0\%, as marked in red color in the tabel. The increases are counted from the performance of the base model. } 
    \setlength{\tabcolsep}{2.0mm}{
    \begin{tabular}{c|c|ccccc}
    \hline
    \hline
    Object Class & \multicolumn{1}{c}{AP↑} & \multicolumn{1}{l}{ATE ↓} & \multicolumn{1}{l}{ASE↓}  & \multicolumn{1}{l}{AOE↓} & \multicolumn{1}{l}{AVE↓} & \multicolumn{1}{l}{AAE↓}\\
    \hline\hline
    car  &  0.495 \textbf{(+5.3\%)} & 0.538 &  0.162 & 0.176 & 0.502 & 0.225 \\
    truck & 0.232 (+4.7\%)&  0.729 & 0.214 &  0.175 & 0.463 & 0.201 \\
    bus & 0.354 \textbf{(+5.0\%)} & 0.681 & 0.201 & 0.108 & 0.890 & 0.321 \\
    trailer & 0.150 (+0.4\%) & 1.036 & 0.241 & 0.364 & 0.350 & 0.162 \\
    cons\_vehicle  & 0.059  (+0.4\%)&  1.105 & 0.488 & 1.269 & 0.092 &  0.397 \\
    pedestrian & 0.275  (+3.4\%)& 0.769 &  0.295 &  0.933 &  0.548 & 0.284  \\
    motorcycle  & 0.306 \textbf{(+6.1\%)} & 0.656 & 0.256 & 0.703 & 0.681 & 0.188 \\   
    bicycle & 0.270  (+2.0\%)& 0.538 & 0.258 &0.967 & 0.229 & 0.007 \\
    traffic\_cone & 0.451 (+3.3\%) & 0.517 & 0.340 & nan & nan & nan \\
    barrier & 0.506  (+3.1\%)& 0.477 & 0.284 & nan & nan & nan \\
    \hline
    \hline
    \end{tabular}%
    }
  \label{tab:per-class}%
\end{table*}%

\begin{table*}[t]
\small
  \centering
  \caption{Detailed records for the performance of MobileNet-V2 student. } 
    \setlength{\tabcolsep}{2.0mm}{
    \begin{tabular}{c|cc|c|cccccc}
    \hline
    \hline
    Model & Image Size & Backbone & \multicolumn{1}{c}{NDS ↑} & \multicolumn{1}{l}{mAP ↑} & \multicolumn{1}{l}{mATE ↓} & \multicolumn{1}{l}{mASE↓}  & \multicolumn{1}{l}{mAOE↓} & \multicolumn{1}{l}{mAVE↓} & \multicolumn{1}{l}{mAAE↓}\\
    \hline \hline

    Base & $256\times704$ & MobileNet-V2 & 0.237 & 0.148 & 0.856 &  0.300 &  1.063 &  0.947 & 0.262  \\

    Student & $256 \times708$ & MobileNet-V2 & 0.275 & 0.165 & 0.822 & 0.276 & 0.835 & 0.856 & 0.269 \\
    
    \hline
    \hline
    \end{tabular}%
    }
  \label{tab:nu_test3}%
\end{table*}%

\subsection{Supplementary description of view-dependent masks selection}
In our experiments, we design camera view-dependent LiDAR masks to guide the distillation process. We obtain the view-dependent masks following the below steps:

Assume we have 6 surround-view images, and for each image, the camera's intrinsics $\mathbf{K}_{3\times3}$, the extrinsics $\mathbf{I^{'}}_{3\times4} = [\mathbf{R^{'}}_{3\times3} ,\mathbf{T{'}}_{3\times1}]$ (from the camera to the car ego), and the LiDAR's extrinsics $\mathbf{I^{''}}_{3\times4} = [\mathbf{R^{''}}_{3\times3} ,\mathbf{T{''}}_{3\times1}]$ (from the LiDAR to the car ego) are all known constants. Then we can transform the image pixels to the ego coordinate:
$$
    Z_c \begin{bmatrix} u & v& 1 \end{bmatrix}^{T} = \mathbf{K^{'}} \times \mathbf{I^{'}} \times \begin{bmatrix} X_e, Y_e, Z_e, 1 \end{bmatrix}^{T}
$$
so, the ego coordinate is obtained:
$$
 \begin{bmatrix} X_e, Y_e, Z_e, 1 \end{bmatrix}^{T} =   (\mathbf{I^{'}})^{-1} \times (\mathbf{K^{'}})^{-1}   Z_c \begin{bmatrix} u & v& 1 \end{bmatrix}^{T} 
$$
as a further step, the ego points can be projected to the LiDAR coordinates:
$$
 \begin{bmatrix} X_l, Y_l, Z_l, 1 \end{bmatrix}^{T} =  \begin{bmatrix} \mathbf{R^{''}} & \mathbf{T^{''}} \\ 0 & 1 \end{bmatrix}^{-1}  \times \begin{bmatrix} X_e, Y_e, Z_e, 1 \end{bmatrix}^{T}
$$

To sum all the steps, the image pixels can be projected to the lidar coordinates via the following functions:
\begin{equation}
    T(\cdot) = \begin{bmatrix} \mathbf{R^{''}} & \mathbf{T^{''}} \\ 0 & 1 \end{bmatrix}^{-1} \times (\mathbf{I^{'}})^{-1} \times (\mathbf{K^{'}})^{-1}   Z_c \begin{bmatrix} u & v& 1 \end{bmatrix}^{T} 
\end{equation}

So given a camera view with depth $\mathbf{V}_i$, $i \in [1,6] \in \mathbb{Z} $, we can obtain the projected pixels in the LiDAR coordinates $P^{i}_{c1l} = T(\mathbf{{V}_i})$. Via a simple BEV Pooling operation $O(\cdot)$, we can obtain the view-dependent mask between the camera and the LiDAR points $P_{LiDAR}$ :
\begin{equation}
    \mathbf{M_i} = O(P_{LiDAR}) \cap  O(P^{i}_{c1l})
\end{equation}
 
\begin{table*}[t]
\small
  \centering
  \caption{Evaluation results on nuScenes validation set for the 3D detection task of the models using CBGS strategy. } 
    \setlength{\tabcolsep}{2.0mm}{
    \begin{tabular}{c|cc|c|cccccc}
    \hline
    \hline
    Model & Image Size & Backbone & \multicolumn{1}{c}{NDS ↑} & \multicolumn{1}{l}{mAP ↑} & \multicolumn{1}{l}{mATE ↓} & \multicolumn{1}{l}{mASE↓}  & \multicolumn{1}{l}{mAOE↓} & \multicolumn{1}{l}{mAVE↓} & \multicolumn{1}{l}{mAAE↓}\\
    \hline \hline

    Base & $256\times704$ & Res-18 & 0.433 & 0.305 & 0.705 &  0.273 &  0.543 &  0.460 & 0.215  \\

    Student & $256 \times708$ & Res-18 & 0.449 & 0.309 & 0.677 & 0.278 & 0.471 & 0.437 & 0.197 \\

    Teacher & $256 \times708$ & Res-101 & 0.478 & 0.343 & 0.646 & 0.276 &0.406 & 0.412 & 0.190 \\   
    \hline
    \hline
    \end{tabular}%
    }
  \label{tab:nu_test2}%
\end{table*}%

\subsection{Supplementary experiment records in the paper}

\textbf{Per-class results of 3D detection}

We supply per-class results of our Res-18 student model. Under LGKD, the best student model achieves 43.4\% NDS score and 31.0 \% mAP on the nuScenes validation set. As recorded in Tabel 1, the student behaves well on predicting cars, barriers and traffic cones, but behaves poorly to predict construction vehicles. We also mark the performance increase of the student model for every class compared with the base model. 

\textbf{Student model results with MobileNet-V2}
In the ablation study, we use MobileNet-V2 as a second choice of the student backbone. And the detailed performance of the student is listed on Tabel 2. The paper's result 14.2\% mAP for the base model is replaced by the true value 14.8\%. 

\textbf{Memory cost}
To reduce the memory cost, we use FP16 by PyTorch-lightning to train our whole framework. Our Res-18 student model takes only 3265 MiB memory testing with the batch-size of 1 on a Nvidia-V100-32G GPU, and we report our student model's advantage over competing baseline techniques in terms of its low memory occupancy and good accuracy.\cite{huang2021bevdet, li2022bevformer, liu2022petr} 

\textbf{Performance in different ranges}
We also test the different-range accuracy of the 3D object detection task for the base model, student model and BEVDepth-R50\cite{li2022bevdepth} model. We calculate $0-10m$, $0-20m$, $0-30m$, $0-40m$ mAP for the three models, and plot the accuracy curve in Figure 1. The mountain-shaped curves have a maximum value between 0 and 20 meters along the distance axis.
\begin{figure}[h]
        \centering
        \includegraphics[width = .9\linewidth]{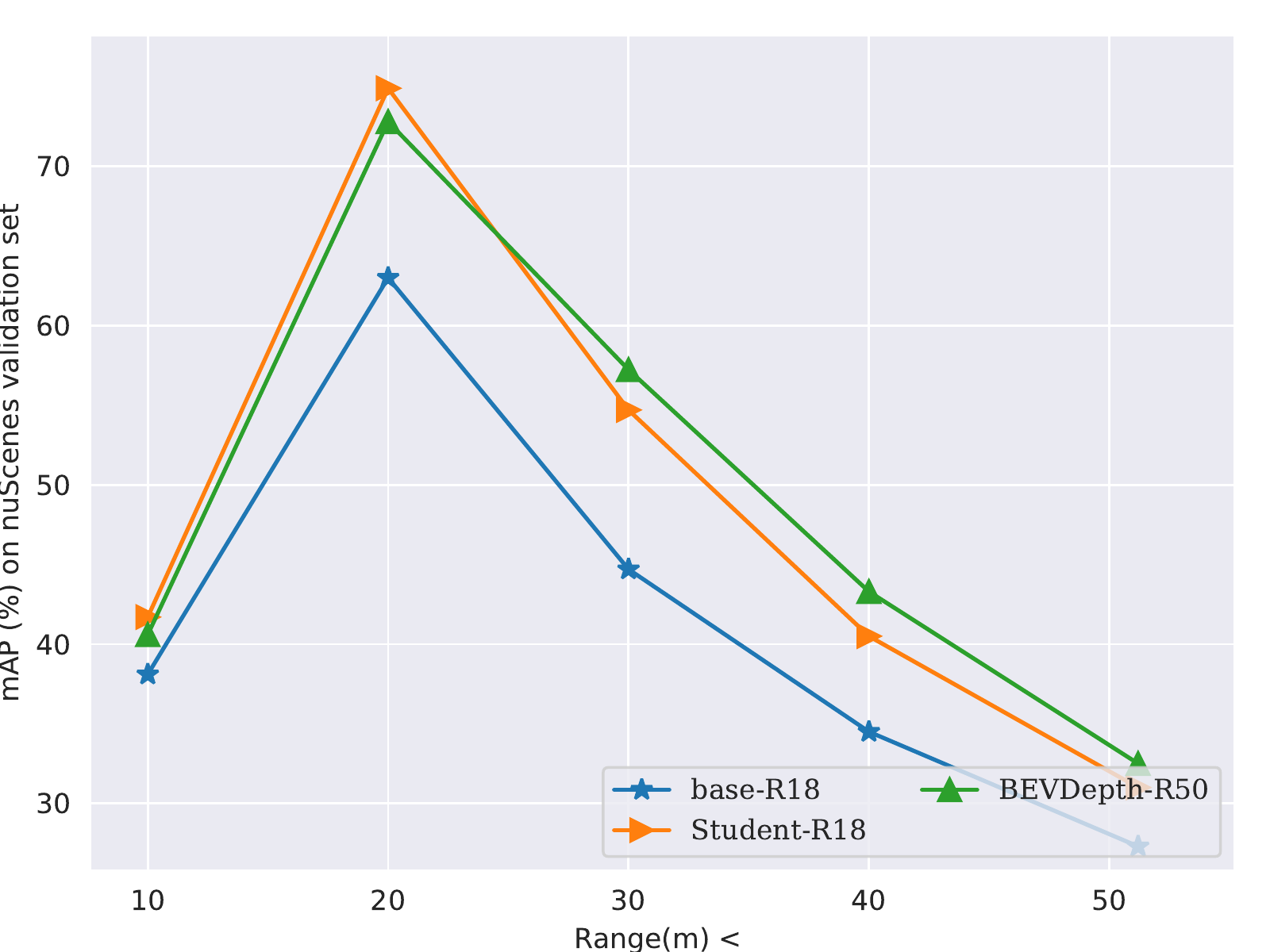}
        \caption{mAP calculated within different ranges.}
\end{figure}

\subsection{Supplementary experimental analysis}
In this section we conduct supplementary experiments to fully support the effectiveness of our LGKD framework. The first experiment is using CBGS\cite{zhu2019class} to enhance the model performance of 3D object detection. The second experiment is evaluating the  accuracy improvements on the BEV segmentation task brought by our LGKD framework.

\subsection{Detection results with using CBGS\cite{zhu2019class}}
Class balanced group strategy\cite{zhu2019class} is a data augmentation strategy designed to generate a more balanced category distribution. The most important step is to create a multi-group head and divide all categories into several groups. We adopt the method in BEVDepth\cite{li2022bevdepth} and use an additional head to aggregate the 3D image features before BEV Pooling. 

With CBGS, we first train the Res-101 teacher model, and then execute the LGKD framework to distill the Res-18 student model from scratch. In Tabel 3, we attach the detailed performance of the student model with CBGS strategy. As shown, the student model achieves 30.9\% on the mAP and 44.9\% on the NDS score. And compared with non-CBGS distillation framework, the student model gains new 0.4\% increase on the mAP and 1.6\% on the NDS score. We also record the learning curve of the student model from epoch 4 to epoch 19 in Figure 1.  

\begin{figure*}[t]
	\centering
	\subcaptionbox{Scene1}{
		\includegraphics[width = .4\linewidth]{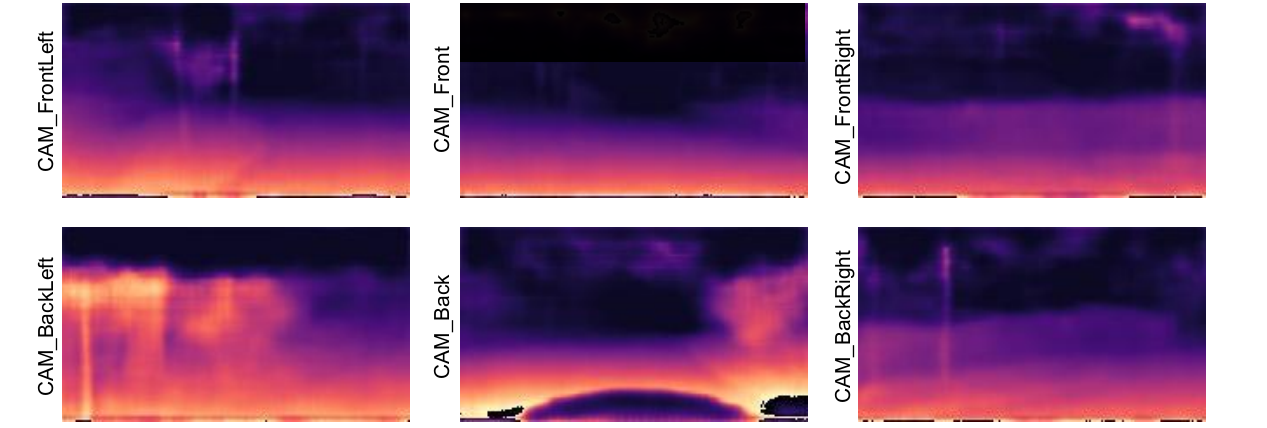}
	}
	\subcaptionbox{Scene2}{
		\centering
		\includegraphics[width = .4\linewidth]{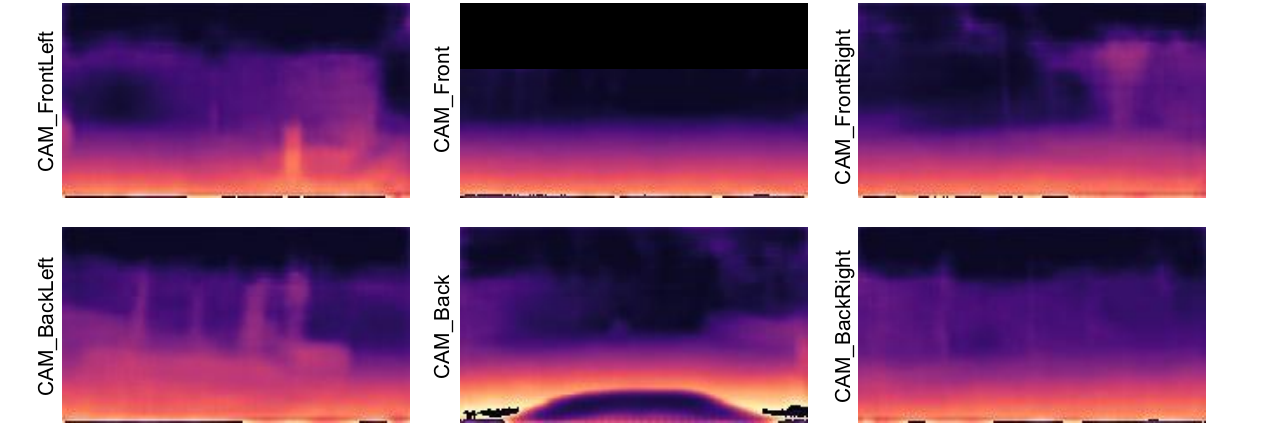}
	}
	\subcaptionbox{Scene3}{
		\includegraphics[width = .4\linewidth]{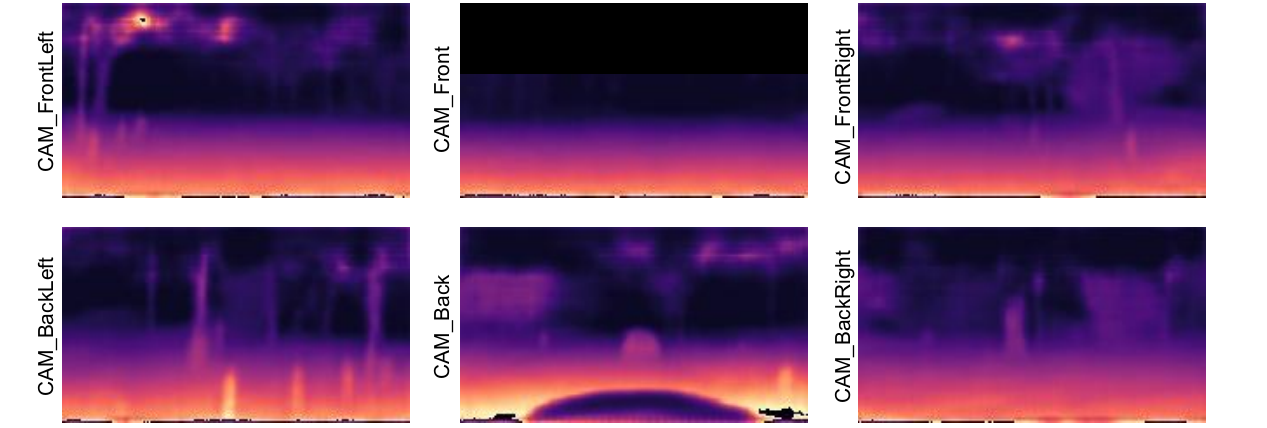}
	}
	\subcaptionbox{Scene4}{
		\includegraphics[width = .4\linewidth]{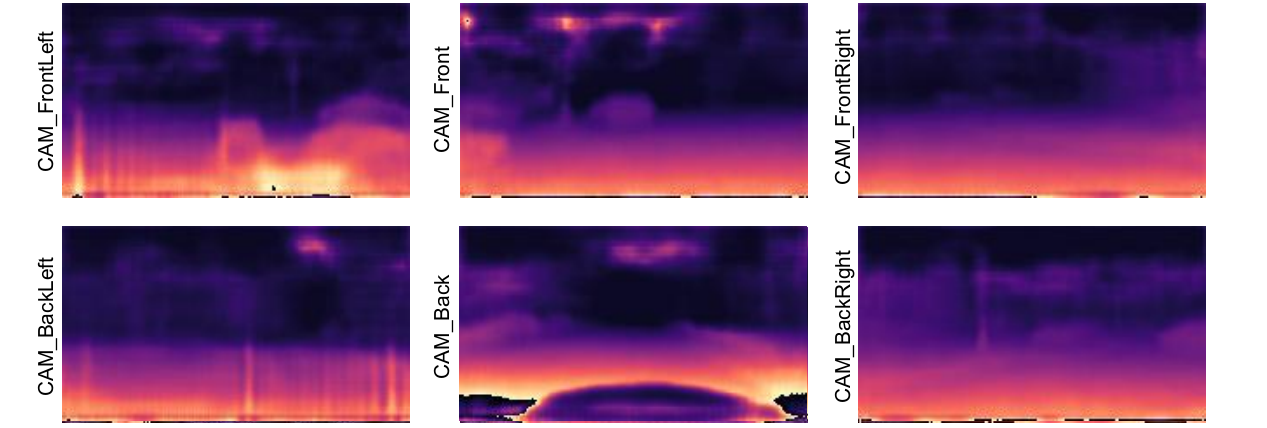}
	}
	\caption{supplementary qualitative results of dense depth estimation from the student model. }
	\label{fig:qual_det1}%
\end{figure*}

\begin{figure}[htbp]
        \centering
        \subcaptionbox{mAP}{
        \includegraphics[width = .45\linewidth]{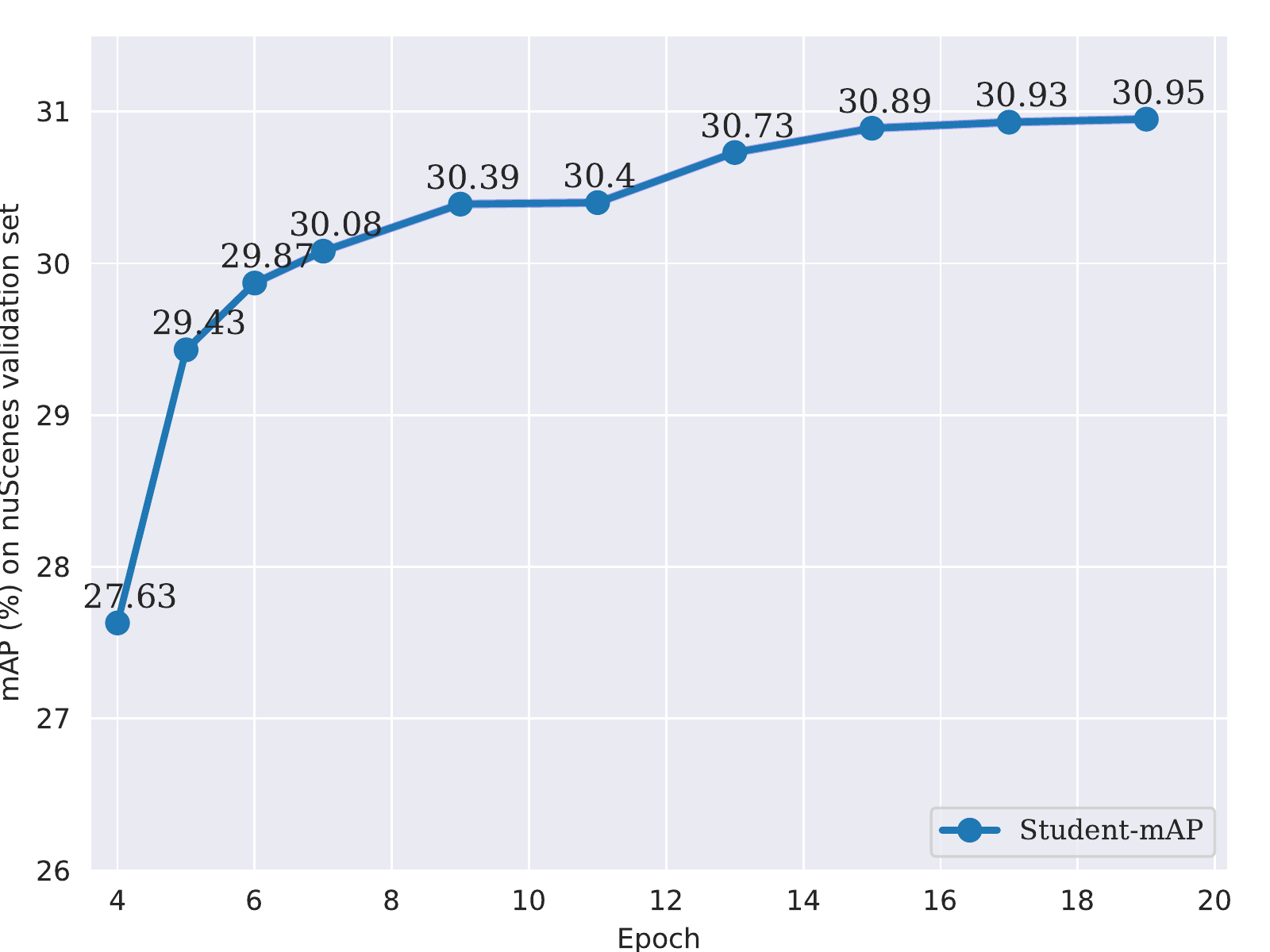}
    }
        \subcaptionbox{NDS}{
        \includegraphics[width = .45\linewidth]{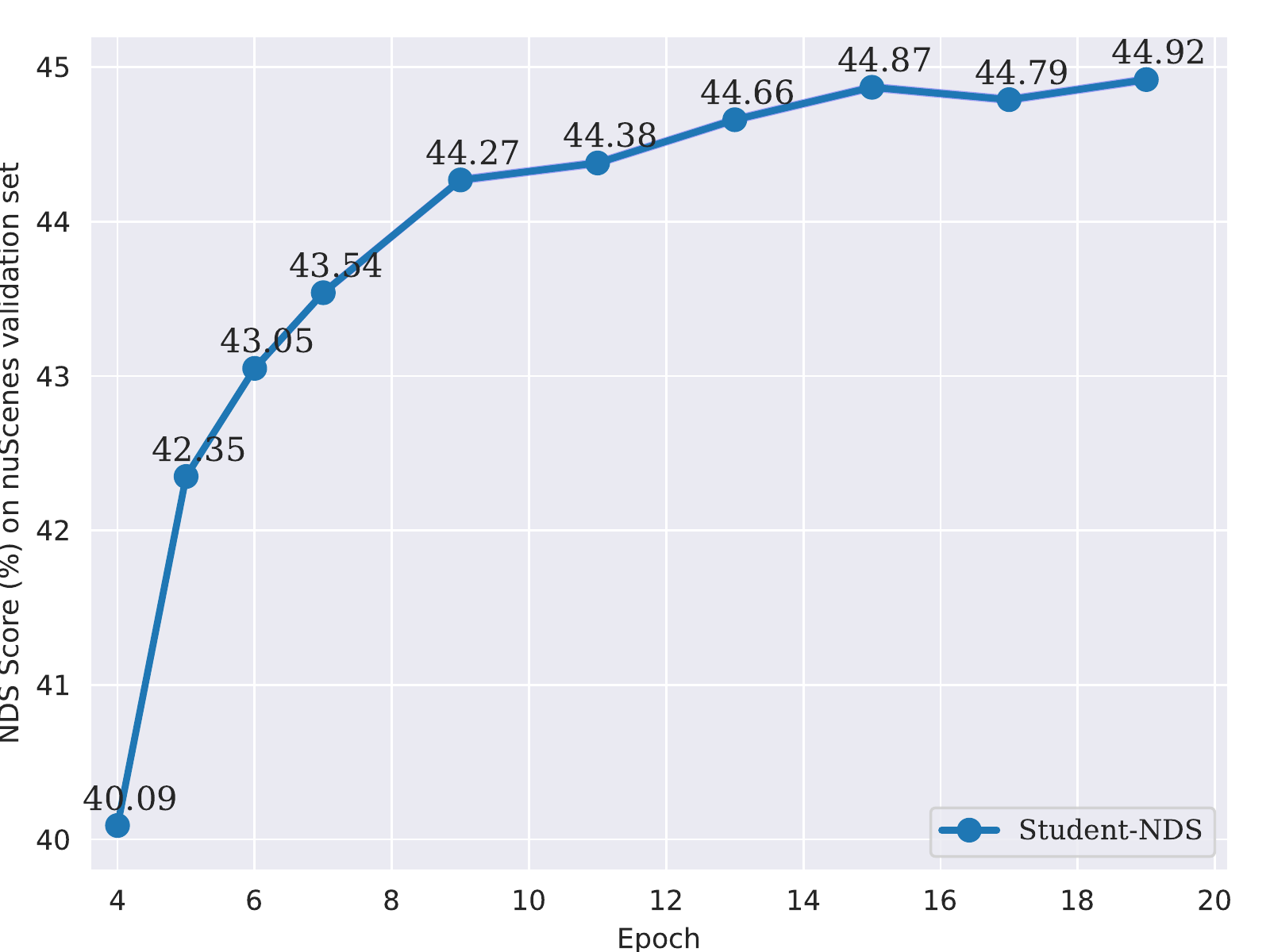}
   }

       \caption{Learning curves of the student model under the LGKD framework with CBGS strategy. We set the max-epoch to 20 for the LGKD distillation process.}
        \label{fig:qual_depth1}%
\end{figure}

\subsection{Results for BEV segmentation}
We also make attempts to demonstrate the effectiveness of our LGKD framework for other BEV tasks. As introduced in Sec. \ref{sec:A}, BEV segmentation aims to understand the surrounding semantics for the multi-view systems under BEV representation. Compared with previous works, we design a simple BEV segmentation framework. We project the bounding boxes of the vehicle instances to the BEV view, and form a binary ground truth. We utilize the segmentation ground truths to supervise the BEV segmentation task using a BCE loss, with a segmentation head to performing vehicle segmentation based on the BEV feature.

Table 4 records the results of the BEV segmentation task. As shown, our LGKD framework can also bring improvement in the BEV segmentation task. The IOU metric increases to 15.2\% from 14.7\%. This further demonstrates that our LGKD framework is able to obtain a more reliable BEV feature for the student model.

\begin{table}[h]

  \centering
  \caption{Experimental records for the BEV segmentation task. "MTL" denotes to "Multi-Task-Learning" of the 3D detection and BEV segmentation tasks.  } 
    \setlength{\tabcolsep}{2.0mm}{
    \begin{tabular}{c|c|c|cc}
    \hline
    \hline
    &  &  \textbf{Segmentation}&  \multicolumn{2}{c}{\textbf{Detection}} \\
    Model & MTL & IOU↑ & \multicolumn{1}{c}{mAP↑}  & \multicolumn{1}{c}{NDS↑} \\
    \hline \hline
    Base & \ding{55} & - & 0.275 & 0.372 \\ 
    Student & \ding{55} & - &  0.305 & 0.425 \\
    \hline
    Base & \checkmark & 0.147 & 0.273 & 0.378 \\ 
    Student & \checkmark & 0.152 &  0.311 & 0.431 \\

    \hline
    \hline
    \end{tabular}%
    }
  \label{tab:nu_test1}%
\end{table}%

\section{Additional visualization results}

\begin{itemize}
    \item 3D Object detection: We supply a demo video to full exhibit the visualization effectiveness of our LGKD framework. The video can be found in our zipped file, named $demo1.mp4$. The demo contains 240 continuous frames in the nuScenes validation set with the frame rate of 12 fps. 
    
    \item Dense depth estimation: We attach more dense depth estimation results in Figure 3. In our submitted paper, we visualize the depth results with color map opencv\_TURBO, in the depth space. In figure 3, we change the color map to opencv\_PLAZMA to visualize the disparity space.

\end{itemize}

\subsection{Code}

will be released at \url{https://github.com/NorthSummer/LGKD}.

\end{document}